\documentclass[10 pt, conference]{ieeeconf}
\IEEEoverridecommandlockouts
\usepackage{graphics} % for pdf, bitmapped graphics files
\usepackage{graphicx} % for pdf, bitmapped graphics files
\usepackage{subfigure}
\usepackage{amsmath} % assumes amsmath package installed
\usepackage{amssymb}  % assumes amsmath package installed
\usepackage{mathtools}
\usepackage{url}
\usepackage{booktabs}
\usepackage[font=small]{caption}
\usepackage{multirow}
\usepackage{afterpage}
\usepackage{url}
\usepackage{color}
\usepackage{threeparttable}

\newcommand{\Cov}{\mathrm{Cov}}

\newcommand{\rn}[1]{%
     \textup{\lowercase\expandafter{\romannumeral#1}}%
}

\title{\LARGE \bf Estimating Metric Poses of Dynamic Objects \\ Using Monocular Visual-Inertial Fusion}
\author{Kejie Qiu$^*$, Tong Qin$^*$, Hongwen Xie$^\dagger$, and Shaojie Shen$^*$
  \thanks{
  $^*$Kejie Qiu, Tong Qin and Shaojie Shen are with the Department of Electronic and Computer Engineering,
  Hong Kong University of Science and Technology, Hong Kong, China.
  {\tt\small kqiuaa@connect.ust.hk, tqinab@connect.ust.hk, eeshaojie@ust.hk}.}
  \thanks{
  $^\dagger$Hongwen Xie is from Pattern Recognition Center of WeChat, Tencent Inc, Beijing, China. {\tt\small hongwenxie@tencent.com}.
  }
  \thanks{
  This work was supported by the WeChat-HKUST Joint Laboratory on Artificial Intelligence Technology (WHAT LAB).
  }}

\begin{document}
\maketitle 

\begin{abstract}
A monocular $3$D object tracking system generally has only up-to-scale pose estimation results without any prior 
knowledge of the tracked object.
In this paper, we propose a novel idea to recover the metric scale of an arbitrary dynamic object by optimizing the trajectory
of the objects in the world frame, without motion assumptions.
By introducing an additional constraint in the time domain, our monocular visual-inertial tracking system can
obtain continuous six degree of freedom ($6$-DoF) pose estimation without scale ambiguity. 
Our method requires neither fixed multi-camera nor depth sensor settings for scale observability, instead, the IMU inside 
the monocular sensing suite provides scale information for both camera itself and the tracked object.
We build the proposed system on top of our monocular visual-inertial system (VINS) to obtain accurate
state estimation of the monocular camera in the world frame. The whole system consists of 
a $2$D object tracker, an object region-based visual bundle adjustment (BA), VINS and 
a correlation analysis-based metric scale estimator.
Experimental comparisons with ground truth demonstrate the tracking accuracy of our $3$D tracking performance
while a mobile augmented reality (AR) demo shows the feasibility of potential applications.
\end{abstract}

\section{Introduction}
\begin{figure}[t]
\begin{center}
  \includegraphics[width=0.8\columnwidth]{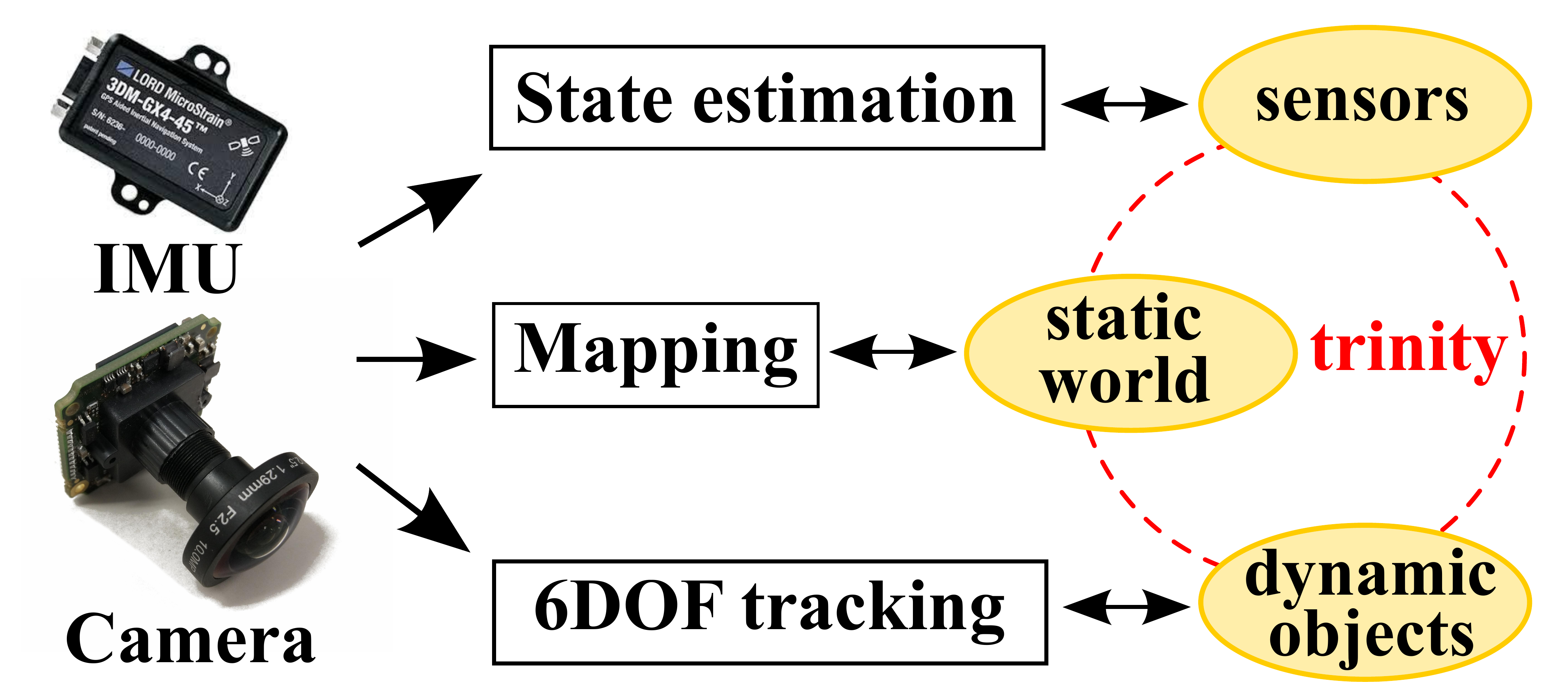}
\end{center}
\caption{The complete perception system based on monocular visual-inertial sensing.}
\label{fig:trinity}
\vspace{-5mm}
\end{figure}

A complete robotic perception system consists of robust state estimation (localization), 
static environment mapping and dynamic objects tracking, through the use of multiple onboard sensors.
Among these diverse sensing options, we are particularly interested in the minimal sensor suite that
consists of only one camera and an IMU, due to its ultra light-weight and low-cost. 
Equipped with state estimation and mapping that comprise a simultaneous localization and mapping (SLAM) system, 
an agent like a quadrotor can execute autonomous navigation within an 
unknown but static environment \cite{lin2017autonomous}.
However, for a complete perception system, dynamic objects in the real world also have to be considered seriously.
Previously, dynamic objects are often regarded as outliers in a SLAM system, and the main consideration is how to make
the autonomous system robust against these outliers. In other words, static environment is a basic assumption. Nevertheless,
a better way is to actively track the dynamic objects regarding $6$-DoF pose estimation,  thus
completing the whole perception system as shown in Fig. \ref{fig:trinity}.
And a lot of benefits will follow if
the specific position and orientation of the dynamic object are known. More robust state estimation, active obstacle avoidance, 
and path planning, manipulation of the objects, even augmented reality effect on the moving objects can become possible.

%Different from the traditional trajectory reconstruction methods, no specific motion assumption about the dynamic object is needed 
%in our system. But 
%Inspired by the way that introducing an additional constraint from the time domain
%for scale estimation, we regard the camera motion and object motion as two isolated signals and recover the metric object
%motion from the perspective of signal processing, say, signal correlation analysis. 
%This process is like the signal decomposition
%problem discussed in the telecommunication community: a time-domain signal segment obtained by a receiver is a combination
%of several isolated signal sources whose signal patterns are pre-designed to satisfy good auto-correlation and cross-correlation
%properties. In this way, the mixed signal could be decomposed into the original signal parts through correlation analysis
%between the mixed signal and known signal patterns. While in our system, the object motion observed from the camera frame 
%corresponds to the mentioned mixed signal (the observed relative motion is a combination of object motion and camera motion), 
%the known signal pattern corresponds to the camera motion in world frame provided by VINS \cite{QinShen17}, and 
%the observability condition is about something similar to the auto-correlation and cross-correlation
%properties, which will be detailed in the methodology part.

Actually, although promising results on state estimate and static environment mapping have been achieved using 
monocular visual-inertial fusion \cite{yang2017real, lin2017autonomous}, six degree of freedom ($6$-DoF) metric 
tracking of dynamic objects remains a 
significant challenge. 
It is obvious that the metric scale or the real depth
value of dynamic objects is of significant importance for various real-time applications.
While the primary difficulty comes from the fact that the metric scale 
is not directly observable from only one camera. 
To handle this, one common solution is to utilize multiple fixed cameras such that the triangulation constraint is still valid even
the objects are moving. Motion capture system such as OptiTrack \footnote{http://optitrack.com/} and 
Vicon \footnote{http://www.vicon.com/} have been a mature 
tracking system for moderately sized application scenarios but the tracked 
objects have to be attached with reflective markers. 

On the other hand, single sensor-based methods using an RGBD sensor \cite{aldoma2013multimodal} or a stereo camera \cite{wang20173d} relax 
the multi-sensor condition at the cost of limited sensing distance. Consequently, these methods can only handle small-scale 
application situations such as vision-based object manipulation. 
%Inspired by the way that introducing an additional constraint from the time domain %\cite{yu2009monocular},
we regard the camera motion and object motion as two isolated signals and recover the metric object
motion from the perspective of signal processing, say, signal correlation analysis. 
This process is like the signal decomposition
problem discussed in the telecommunication community.
Thus, we propose to estimate the metric scale
by analyzing the temporally statistical relationship between the camera motion and the recovered object motion.

%We identify our contributions as follows:
%\begin{itemize}
%  \item We propose a practical $6$-DoF object tracking system using a monocular visual-inertial sensor suite, and 
%      without any prior $3$D information of the object.
%  \item We complete the 3D tracking system by estimating the metric scale of the tracked object based on
%      motion correlation analysis and observability condition check.
%  \item We implement the $3$D tracking system to verify the tracking performance and demonstrate the AR application potential.
%\end{itemize}

The rest of the paper is structured as follows.
Section~\ref{sec:related} introduces relevant work on $6$-DoF object pose estimation and fundamental work that our 
system relies on.
In section~\ref{sec:overview}, our system structure and major frame relationships are illustrated briefly.
Section~\ref{sec:methodology} describes how the objective functions are formulated and solved 
for up-to-scale $3$D tracker and metric scale estimation.
Section~\ref{sec:experiments} gives the experimental results with comparison against ground truth.
Conclusion and directions for future work are presented in Section~\ref{sec:conclusion}.

\section{Related work}
\label{sec:related}
Estimating the $6$-DoF pose of a dynamic object using multiple fixed cameras is a well-studied formulation.
Besides the commercial active-tracking system with specialized markers, passive-tracking systems for sports 
scenes analyzing and traffic surveillance were also proposed since the targets are usually non-cooperative in these cases.
A UAV trajectory estimation system using flight dynamics as a prior was proposed based on fixed ground cameras,
assuming that all the camera videos are synchronized \cite{rozantsev2016flight}.
\cite{vo2016spatiotemporal} proposed a spatiotemporal Bundle Adjustment framework to simultaneously estimate the temporal alignment
between cameras. However, all of these methods are limited by the fixed-camera setting.

Oppositely, few methods adopt a monocular camera framework 
to recover the whole $6$-DoF pose of dynamic objects, since the tracking results may lack metric scale unless 
the tracked target is studied or modeled carefully in advance. 
However, solving this problem only using single camera has attracted even more attention thanks to
its ultra light-weight, strong adaption, and synchronization-free features compared to the fixed 
multi-camera setting. 
Model-based tracking is one branch of monocular-based methods.
Given a set of known $2$D-to-$3$D correspondences, the relative pose can easily be solved by using Perspective-n-Point (PnP) \cite{garrido2014automatic} or view alignment using edge features \cite{qiu2017model, qiu2017model2}.
For instance, ARUCO tag \cite{Aruco2014} made use of pre-designed tags as the known $3$D information to mark tracked targets, 
and calculate the camera pose in terms of the tags.
Instead of using pre-designed landmarks, discriminative feature points such as BRISK \cite{leutenegger2011brisk} are also practical for 2D-to-3D matching. 
In addition to sparse point features, dense method \cite{brachmann2016uncertainty} and edge-based method \cite{choi20123d} are good alternatives for
textureless object tracking. All the methods work well as long as the $3$D model of the tracked object was 
carefully modeled in advance.
Another significant branch is learning-based tracking \cite{zeng2017multi}, 
Georgios Pavlakos et al. proposed an efficient convolutional network to 
first locate reliable pre-defined semantic keypoints and then estimate the $6$-DoF pose with only a monocular camera \cite{pavlakos20176}.
However, a faked scenario can destroy learning-based methods easily especially in AR applications. 
For example, a real car and a car model can fool 
learning-based methods for lack of metric scale measurement.
%cannot handle arbitrary objects' tracking
%since both
%of them are categorized as prior-based methods, the model of the tracked object has to be known
%or the object has to be trained in advance.

In order to estimate the metric object scale, the camera pose has to be known in advance such that the object 
pose computed from the region-based BA can be projected from the camera frame to the world frame.
And we make use of our visual-inertial system (VINS) for accurate and robust camera pose estimation in the world frame,
interested readers are referred to our previous work on state estimation with visual-inertial fusion \cite{yang2016monocular, QinShen17}.
Also, our method is designed to handle $6$-DoF tracking of an arbitrary rigid object by using an object region-based 
visual bundle adjustment (BA) with online scale estimation. 
In order to obtain accurate object regions for region-based visual BA, a robust $2$D tracker is needed. 
The $2$D tracking problem has already been well discussed and lots of classic tracking methods such as CMT \cite{nebehay2015clustering},
STRUCK \cite{hare2016struck} are robust enough for most cases, resulting in sequential bounding boxes. 
Given accurate object areas on the sequentially images, a region-based BA could be 
used to get up-to-scale relative $6$-DoF pose. 

%Finally, the metric scale estimation is inspired by the methods 
%used in solving a problem called trajectory reconstruction.
%Trajectory reconstruction is the problem of solving point trajectory from a monocular image sequence.
%The tracked object is regarded as a point such that the rotation of the object cannot be recovered. These methods try 
%to estimate multiple object depths in all image frames instead of an intrinsic scale of the object.
%Shai Avidan \cite{avidan2000trajectory} solved straight-line and planar conic motions.
%\cite{yu2009monocular, kaminski2004general} extended it to general motion cases with polynomial fitting, 
%in which the trajectory is represented by a linear combination of polynomial bases.
%While \cite{park20103d} made use discrete cosine transform (DCT) basis as the object trajectory bases.
%These fitting-based methods utilize the universal smoothness constraint but highly rely on 
%fitting model selection for good performance. The rotation of the object is not estimated either.

\begin{figure}[t]
\begin{center}
  \includegraphics[width=0.8\columnwidth]{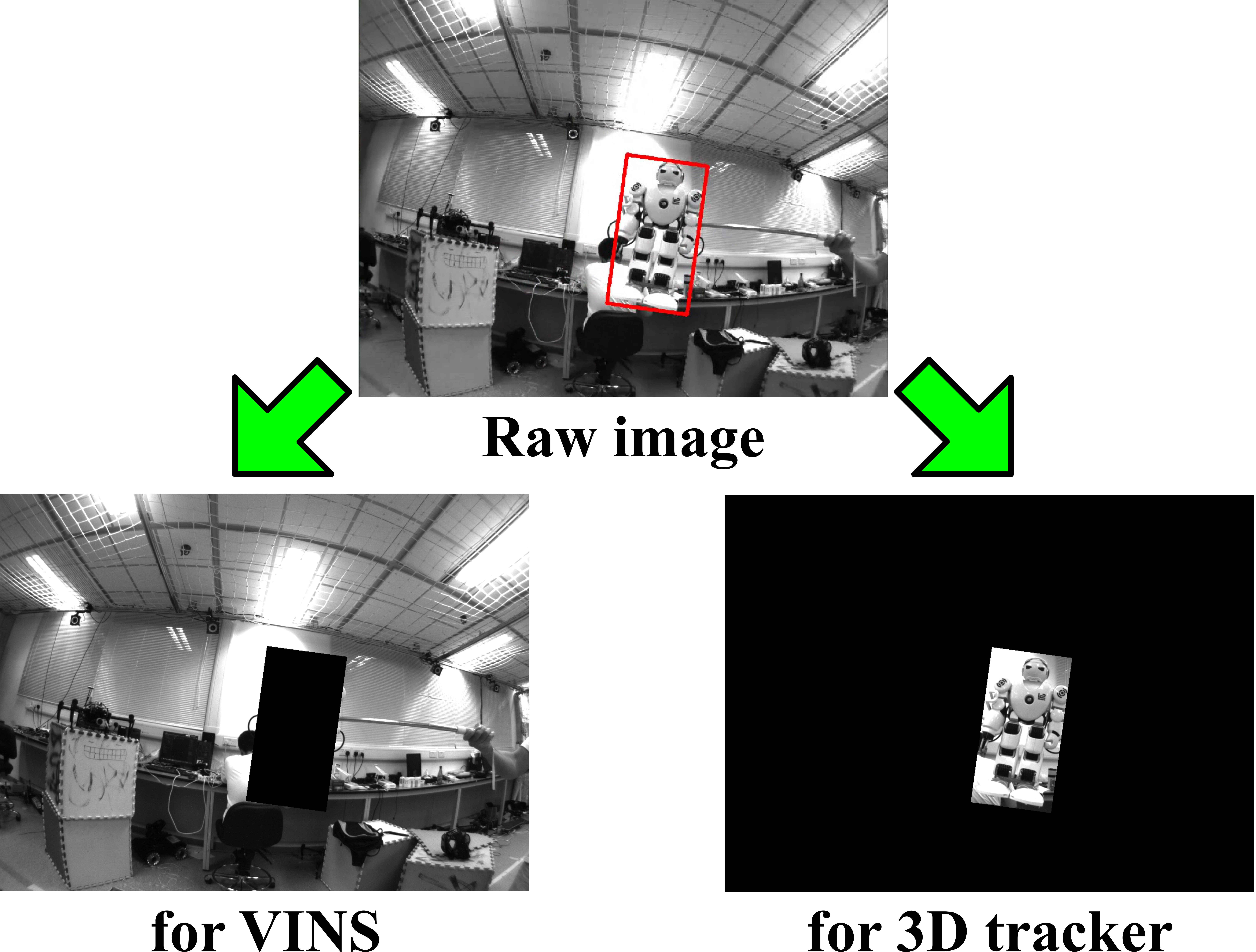}
\end{center}
\caption{$2$D object tracking results with bounding boxes.}
\label{fig:bounding_box}
\end{figure}

\begin{figure}[t]
\begin{center}
  \includegraphics[width=0.8\columnwidth]{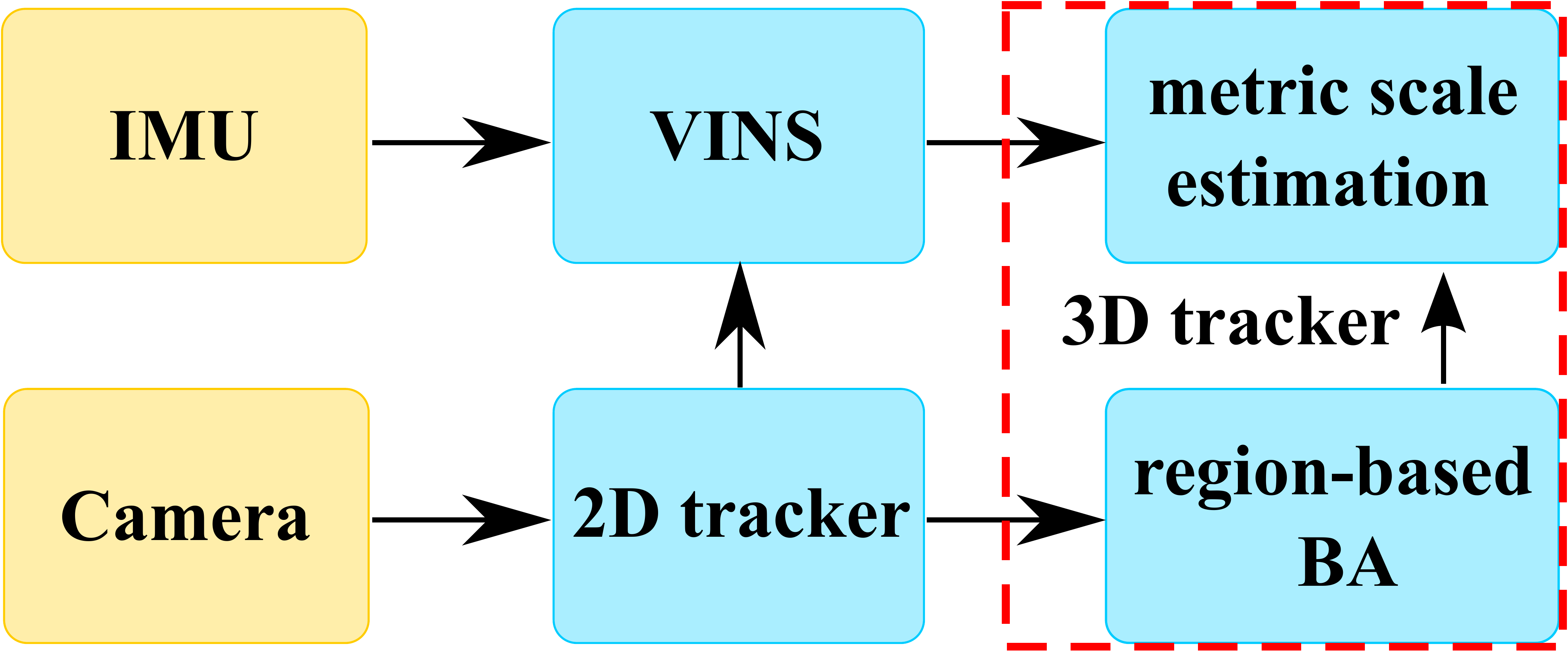}
\end{center}
\caption{Overall system structure.}
\label{fig:system}
\vspace{-5mm}
\end{figure}

\begin{figure}[t]
\begin{center}
  \includegraphics[width=0.8\columnwidth]{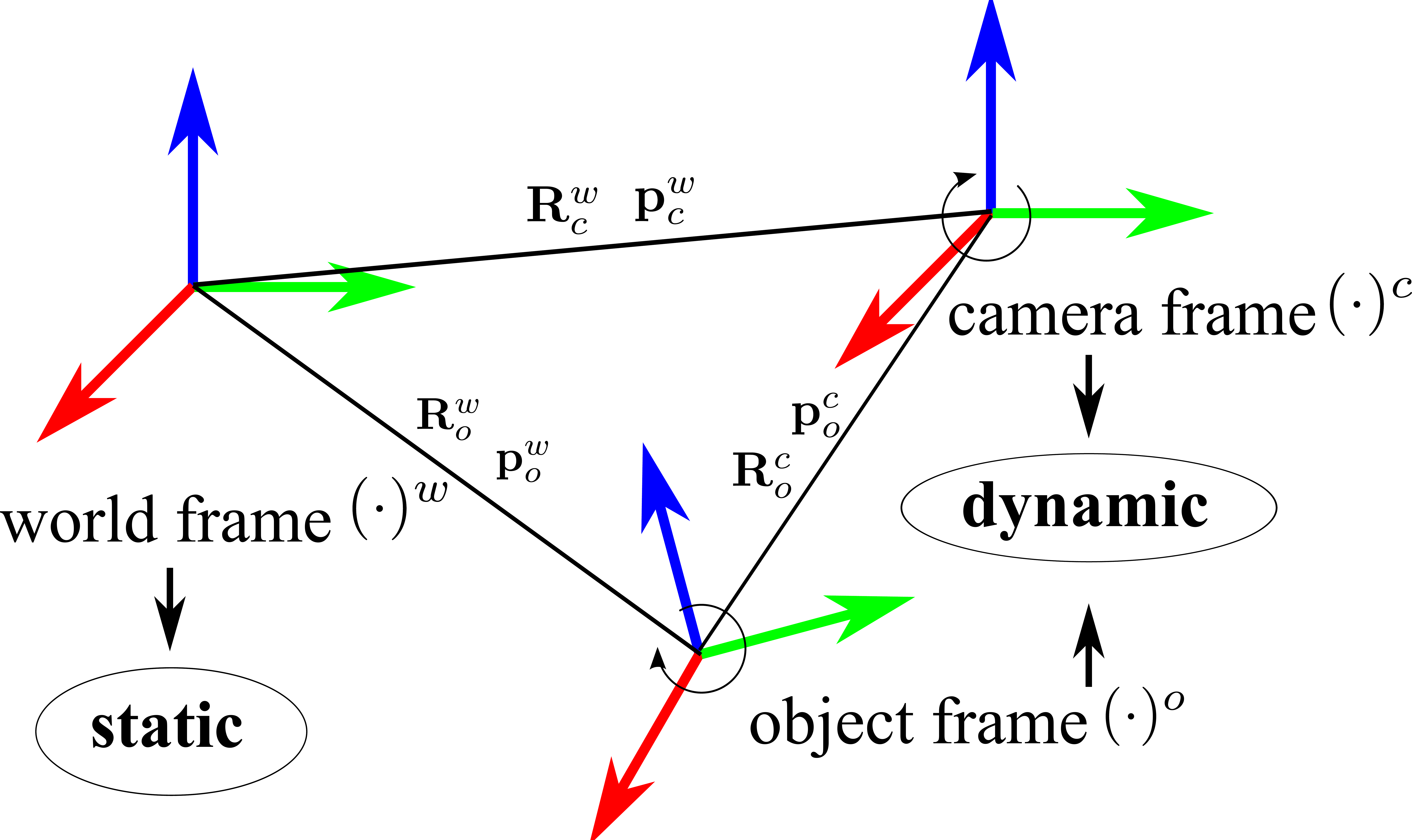}
\end{center}
\caption{Coordinate frames in the system.}
\label{fig:frames}
\vspace{-5mm}
\end{figure}

\section{System Overview}
\label{sec:overview}
The captured images are first processed by $2$D object tracking, resulting in object regions represented by $2$D bounding boxes 
as shown in Fig. \ref{fig:bounding_box}.
VINS \cite{QinShen17} takes both IMU and masked images for tightly-coupled camera pose estimation in the world frame. 
%Note that we do not need to know which type of tracked object is. Nor do we need to have any prior motion assumptions
%about the object. For example, we can track and recover the $3$D motion for both a real car and a car model, although
%they have different sizes but similar visual appearance.
Given consecutive image regions of the object, a region-based visual BA is running for up-to-scale object pose estimation.
Finally, the camera pose in the world frame and the object pose in the camera frame are collected together for 
metric scale estimation. Once the metric scale estimation obtained, the metric object poses in the world
frame is recovered. 
%Since the object is regarded as the reference frame in the region-based 
%BA, the tracked object must be a rigid one, which is the basic assumption of our tracking system.
The overall system structure is shown in Fig. \ref{fig:system}. 
The combination of the region-based BA and metric scale estimation is also called $3$D tracker,
while up-to-scale $3$D tracker particularly refers to the region-based BA.
In this paper, we assume that a $2$D object 
detector is provided for $2$D tracking initialization and focus on region-based visual bundle adjustment and 
online metric scale estimation. %Both VINS and region-based BA are formulated as sliding window-based nonlinear optimizations. 
%A unified representation of the sliding window structure is shown in Fig. \ref{fig:sliding_window}.

The relationship between relevant frames is shown in Fig. \ref{fig:frames}, where $(\cdot)^w$ is the world frame, $(\cdot)^c$ 
the camera frame, and $(\cdot)^o$ is the object frame. The IMU frame is ignored since it only constrains the metric scale of 
VINS.
$\mathbf{p}_y^x$ and $\mathbf{R}_y^x$ are the 3D translation and rotation of frame $(\cdot)^y$ respectively with respect to frame $(\cdot)^x$.
The camera pose in the world provided by VINS is represented by $\mathbf{R}_c^w$ and $\mathbf{p}_c^w$,
and $\bar{(\cdot)}$ denotes the up-to-scale pose results. Thus $\mathbf{R}_o^c$ and $\bar{\mathbf{p}}_o^c$ together denotes
the relative transformation of the object frame in terms of the camera frame
provided by the region-based visual BA module. What the metric scale estimation module estimates is the scale ratio $s$
that makes the up-to-scale position to scaled one ($\mathbf{p}_o^c=s\cdot\bar{\mathbf{p}}_o^c$).
In the end, we use $\mathbf{R}_o^w$ and $\mathbf{p}_o^w$ together to denote the global transformation from the world frame to the object frame.
The detailed representation of the global object pose can be found in Section~\ref{sec:methodology}.

%\begin{figure*}[t]
%\begin{center}
%  \includegraphics[width=1.8\columnwidth]{fig/sliding_window.pdf}
%\end{center}
%\caption{A unified illustration of the sliding window structures for both VINS and region-based visual bundle adjustment in the world frame.
%Five camera states together with four fixed environment features and three dynamic object features are drawn.}
%\label{fig:sliding_window}
%\end{figure*}

\section{Methodology}
\label{sec:methodology}
\subsection{Up-to-scale 3D tracker}
The $3$D tracker we design is based on a region-based visual bundle adjustment, 
we construct a purely vision-based 
graph optimization for 3D pose tracking of general objects. 
In this way, any rigid objects with arbitrary shapes could be tracked since no assumption about the object shape is needed.
Note that this bundle adjustment is performed with respect to the object frame, that is,
the $3$D tracker is to estimate the camera motion w.r.t the object frame, which is different from the visual
BA defined in VINS with estimating the camera motion w.r.t the world frame.
Since both camera frame and object frame are dynamic, either camera motion or object motion can 
cause the relative motion detected by the  
region-based visual BA. That is to say, the estimated motion of the $3$D tracker is a 
compound motion coupled with two independent physical motions.

The full state of a sliding window with $N$ image frames and $M$ object features 
is defined as follows:
\begin{equation}
\begin{split}
\mathcal{X}    &= \left [ \mathbf{x}_0,\,\mathbf{x}_{1},\, \cdots \,\mathbf{x}_{N-1},\, \mu_0,\,\mu_{1},\, \cdots \,\mu_{M-1} \right ] \\
    \mathbf{x}_k   &= \left [ \bar{\mathbf{p}}^{c_k}_{o_k},\,\mathbf{q}^{c_k}_{o_k} \right ], k\in [0,N-1],
\end{split}
\end{equation}

where the $k$-th camera state consists of the up-to-scale position $\bar{\mathbf{p}}^{c_k}_{o_k}$, and 
orientation $\mathbf{q}^{c_k}_{o_k}$ of camera frame $c_k$ with respect
to object frame $o_k$. 3D features are parameterized by their inverse depth $\mu$ when first observed in camera frame. The bundle adjustment 
can be formulated as a nonlinear least-square problem, 

\begin{equation}
\label{eq:nonlinear_cost_function}
\begin{aligned}
\min_{\mathcal{X}} 
\sum_{(l,j) \in \mathcal{C}} \left \| \mathbf{r}_{\mathcal{C}}(\hat{\mathbf{z}}^{c_j}_l ,\, \mathcal{X}) \right \|_{2}^2
,
\end{aligned}                    
\end{equation}
where $\mathbf{r}_{\mathcal{C}}(\hat{\mathbf{z}}^{c_j}_l ,\, \mathcal{X})$ is the 
nonlinear residual function of visual measurements.
%$\| \cdot \|$ is the Mahalanobis distance weighted by covariance $P$.
The vector $\mathbf{z}^{c_j}_l$
results from the procedure where the $l$th feature is projected to camera frame $c_j$.
And the visual residual is defined as the sum of the reprojection error between the projected
3D features and the observed $2$D features.
Since this formulation takes the object 
regions of the captured images as input, the object is used as the reference frame for 
camera pose estimation. The front end of the region-based 
visual BA for feature association adopts the same corner features as that used in VINS, which
could be replaced by other feature-based methods, such as edge-based and dense methods to handle featureless objects.

\begin{figure}[t]
\begin{center}
  \includegraphics[width=0.8\columnwidth]{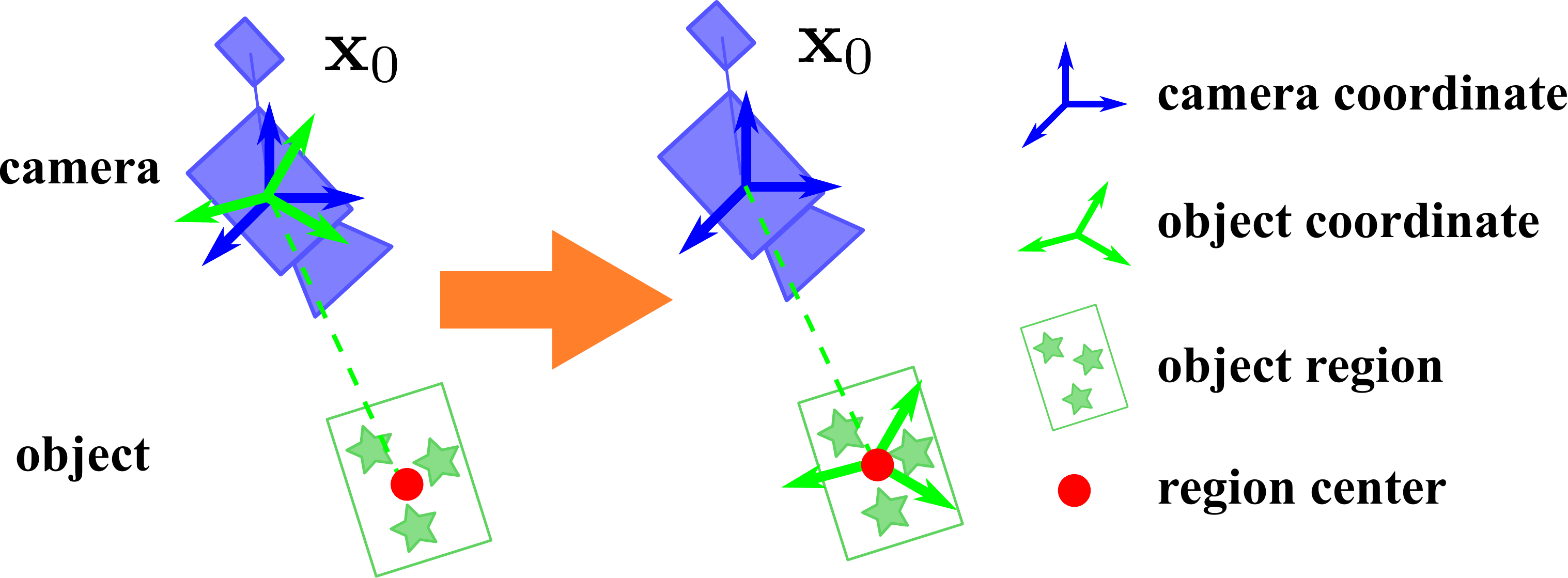}
\end{center}
\caption{Object coordinate initialization process.}
\label{fig:obj_init}
\vspace{-5mm}
\end{figure}

The direct optimization results are camera transformations in terms of the object frame, 
while what we need for $3$D object 
tracking is the object pose in the camera frame. Thus, the object coordinate has to
be located around the object itself, which is not guaranteed by
a regular visual BA. 
%Because we set the initial camera position in terms of the object frame
%as zero like most bundle adjustment methods do,
%this coordinate initialization way is no problem for 
%visual SLAM systems since the world coordinate can be anywhere, but not for 3D object tracking
%where the object coordinate is dynamic and the original point marks the object location.
To solve this problem, we further modify the initial poses by estimating the 
initial object pose in the camera frame. As shown in Fig. \ref{fig:obj_init}, 
we move the object coordinate from the initial position where the initial camera optical center locates 
to the object surface along the direction denoted by the $2$D object region center,
and normalize the object depth by scaling
the object point cloud simultaneously. In fact, the object coordinate can be 
anywhere around the point cloud on the object, once it's determined in the initialization process, 
it will be continuously tracked in the bundle adjustment framework.

\subsection{Metric scale estimation}
Before doing metric scale estimation, we already have accurate enough camera poses in the world frame, and
up-to-scale object poses in the camera frame. 
%With a metric scale estimate, we can recover
%the object poses in the world frame through simple rigid body transformation.
Fundamentally, the metric scale of a tracked rigid object is unobservable based on a monocular camera.
Consider an extreme case that the monocular sensor suite is always static when tracking the object, 
the IMU that involves scale information will have no contribution to the object scale recovery. However, the metric scale
estimation problem becomes conditionally solvable
if the camera motion and the object motion follow a specific condition, or scale observability condition.
%Different from depth estimation that should be running all the time, 
Since the intrinsic scale estimation
could work with the region-based bundle to provide continuous metric pose tracking once the scale is estimated,
so we can accumulate a period of observations, waiting for an opportunity that the designed
observability condition is satisfied.

The metric scale estimation problem in this paper is solved from the perspective of statistics and probability.
As we know the direct observation from the region-based bundle adjustment is a compound motion of the object
motion and camera motion, while the object motion and camera motion are supposed to be independent since 
they have no real physical connections. Thus the scale could be estimated based on the 
statistical independence of the object motion and camera motion in the world frame.

We start by analyzing the scale factor ($s$) we want to estimate, it is actually a scale ratio of the real metric scale
over the inherent scale exists in the region-based BA ($\mathbf{p}_{o_t}^{c_t}=s\cdot\bar{\mathbf{p}}_{o_t}^{c_t}$).
It is time-invariant if the region-based BA is accurate enough since it reflects the intrinsic scale of the tracked object.
For the same reason, the intrinsic metric scale estimator needs not to perform for each input image.
We call this scale ratio as scale in the following content for simplicity.
Accordingly, the estimated scale $\hat{s}$ could be represented as the sum of the true scale $s$ and 
the estimation error $\Delta s$:
\begin{equation}
\label{equ:object_motion_world2}
\hat{s}=s+\Delta s.
\end{equation}

Recall the recovered object trajectory:
\begin{equation}
\label{equ:object_motion_world3}
\begin{aligned}
    \hat{\mathbf{p}}_{o_t}^w&=\hat{s}\cdot\mathbf{R}_{c_t}^w\bar{\mathbf{p}}_{o_t}^{c_t} + \mathbf{p}_{c_t}^w
                            =\hat{s}\cdot \mathbf{p}_{d_t}^{w} + \mathbf{p}_{c_t}^w,
                            %&=(s+\Delta s)\cdot \mathbf{p}_{d_t}^{w}  + \mathbf{p}_{c_t}^w\\
                            %&=\mathbf{p}_{o_t}^w+\Delta s\cdot \mathbf{p}_{d_t}^{w} \\
                            %&= \mathbf{p}_{o_t}^w+\frac{\Delta s}{s}(\mathbf{p}_{o_t}^{w}-\mathbf{p}_{c_t}^{w})\\
                            %&=(1+\frac{\Delta s}{s})\mathbf{p}_{o_t}^w-\frac{\Delta s}{s}\mathbf{p}_{c_t}^{w},
\end{aligned}
\end{equation}
where 
\begin{equation}
\label{equ:pd}
\mathbf{p}_{d_t}^{w}=\mathbf{R}_{c_t}^w\bar{\mathbf{p}}_{o_t}^{c_t}.
\end{equation}

Particularly, when $\hat{s}$ equals to $s$:
\begin{equation}
\label{equ:object_motion_world4}
\begin{aligned}
    \mathbf{p}_{o_t}^w&=s\cdot \mathbf{p}_{d_t}^{w} + \mathbf{p}_{c_t}^w.\\
\end{aligned}
\end{equation}

Thus:

\small
\begin{equation}
\label{equ:object_motion_world5}
\begin{aligned}
    \hat{\mathbf{p}}_{o_t}^w&=\hat{s}\cdot \mathbf{p}_{d_t}^{w} + \mathbf{p}_{c_t}^w
                            =(s+\Delta s)\cdot \mathbf{p}_{d_t}^{w}  + \mathbf{p}_{c_t}^w\\
                            &=\mathbf{p}_{o_t}^w+\Delta s\cdot \mathbf{p}_{d_t}^{w} 
                            = \mathbf{p}_{o_t}^w+\frac{\Delta s}{s}(\mathbf{p}_{o_t}^{w}-\mathbf{p}_{c_t}^{w})\\
                            &=(1+\frac{\Delta s}{s})\mathbf{p}_{o_t}^w-\frac{\Delta s}{s}\mathbf{p}_{c_t}^{w},
\end{aligned}
\end{equation}
\normalsize
from the recovered object trajectory representation we can see that uncorrected scale estimation results in 
insufficient motion decomposition, which means the recovered object motion has some correlation with the camera 
motion. Thus the optimal scale could be estimated by correlation analysis between the 
recovered object motion and the corresponding camera motion.
If the object motion and camera motion satisfies 
a certain observability condition within a time domain, we argue that an incorrect metric scale 
estimate, no matter larger or smaller than the true scale value, 
corresponds to a kind of salient correlation relationship 
between the recovered object motion and the camera motion. 
Thus a good scale estimate ($\Delta s \to 0$)
could be obtained by minimizing
the correlation between the recovered object motion and the camera motion. 
Thus, the key objective function formulation is all about how to quantize the correlation properly.

We first model the recovered object motion and camera motion as two random variables $\mathbf{m}_o$ 
and $\mathbf{m}_c$, and 
define another random variable $\mathbf{m}_d$ to denote the up-to-scale motion difference, here we adopt a certain 
order derivative of a motion trajectory for motion representation:
\normalsize
\small
\begin{equation}
\label{equ:RV}
\begin{aligned}
    \mathbf{m}_c=&[m_c^x, m_c^y, m_c^z]^T=\frac{\partial^n \mathbf{p}_{c}^w(t)}{\partial t^n},\\
    \mathbf{m}_d=&[m_d^x, m_d^y, m_d^z]^T=\frac{\partial^n \mathbf{p}_{d}^w(t)}{\partial t^n},\\
    \hat{\mathbf{m}}_{o}=&\hat{\mathbf{m}}_o(\hat{s}) =[\hat{m}_{o}^x, \hat{m}_{o}^y, \hat{m}_{o}^z]^T=\frac{\partial^n \hat{\mathbf{p}}_{o}^w(t)}{\partial t^n}\\ 
    =&\hat{s}\mathbf{m}_d + \mathbf{m}_c
    =(1+\frac{\Delta s}{s})\mathbf{m}_o - \frac{\Delta s}{s}\mathbf{m}_c.
\end{aligned}
\end{equation}
\normalsize
For example, $n=2$ corresponds to the acceleration, and we directly approximate the motions from the raw data:
\normalsize
\small
\begin{equation}
\label{equ:dynamics_approximation}
\begin{aligned}
    \mathbf{m}_c&=\frac{\partial^2 \mathbf{p}_{c}^w(t)}{\partial t^2}  \approx \frac{\mathbf{p}^w_{c_{t+2\Delta t}}-2\mathbf{p}^w_{c_{t+\Delta t}}+\mathbf{p}^w_{c_{t}}}{\Delta t^2},\\
    \mathbf{m}_d&=\frac{\partial^2 \mathbf{p}_{d}^w(t)}{\partial t^2}  \approx \frac{\mathbf{p}^w_{d_{t+2\Delta t}}-2\mathbf{p}^w_{d_{t+\Delta t}}+\mathbf{p}^w_{d_{t}}}{\Delta t^2}.\\
    %\mathbf{m}_d=\mathbf{p}_d^w(t)\approx p_d(i+2)-2p_d(i+1)+p_d(i)\\
\end{aligned}
\end{equation}
\normalsize

We will further design an observability condition later such that the quantitative correlation value
between $\hat{\mathbf{m}}_{o}$ and $\mathbf{m}_c$ could be 
minimized for a good scale estimation $\hat{s}$.
Take a look at this covariance matrix:
\begin{equation}
\label{equ:cov_oc}
\begin{aligned}
    \Cov&(\hat{\mathbf{m}}_{o},\mathbf{m}_c)\\
    =& E[(\hat{\mathbf{m}}_{o} - E[\hat{\mathbf{m}}_{o}])(\mathbf{m}_c - E[\mathbf{m}_c])^T   ]\\
    =& E[ (\hat{s}\mathbf{m}_d+\mathbf{m}_c -E[\hat{s}\mathbf{m}_d+\mathbf{m}_c])(\mathbf{m}_c - E[\mathbf{m}_c])^T]\\
    =& \hat{s}\Cov(\mathbf{m}_d,\mathbf{m}_c)+\Cov(\mathbf{m}_c, \mathbf{m}_c).\\
\end{aligned}
\end{equation}

Particularly, when $\hat{s}$ equals to $s$:
\begin{equation}
\label{equ:cov_oc_particular}
\begin{aligned}
    \Cov(\mathbf{m}_{o},\mathbf{m}_c)
    =s\Cov(\mathbf{m}_d,\mathbf{m}_c)+\Cov(\mathbf{m}_c, \mathbf{m}_c).\\
\end{aligned}
\end{equation}

All the population covariances could be estimated from the latest $N_o$ observations 
($N_o$ is a big enough but a limited number for bounded computation complexity consideration).
\normalsize
\small
\begin{equation}
\label{equ:objective_approx}
\begin{aligned}
    &\Cov(\mathbf{m}_d,\mathbf{m}_c) \approx \frac{1}{N_o-1} \sum\limits_{k=1}^{N_o} (\mathbf{m}_d^{k} - \hat{\mathbf{m}}_d) (\mathbf{m}_c^{k} - \hat{\mathbf{m}}_c)^T,\\
    &\Cov(\mathbf{m}_c,\mathbf{m}_c) \approx \frac{1}{N_o-1} \sum\limits_{k=1}^{N_o} (\mathbf{m}_c^{k} - \hat{\mathbf{m}}_c ) (\mathbf{m}_c^{k} - \hat{\mathbf{m}}_c )^T,\\
    &\hat{\mathbf{m}}_d = \frac{1}{N_o} \sum\limits_{k=1}^{N_o} \mathbf{m}_d^{k},
    ~\hat{\mathbf{m}}_c = \frac{1}{N_o} \sum\limits_{k=1}^{N_o} \mathbf{m}_c^{k}.
\end{aligned}
\end{equation}
\normalsize

The standard correlation measurement between two multivariate random variables is to use
trace correlation \cite{hooper1959simultaneous}. Instead, we utilize a simplified measurement in terms of individual covariances
such that a closed form scale estimation could be derived.
The objective function $f(\hat{s})$ is formulated as the sum of all the covariance squares of Equation \ref{equ:cov_oc}:

\small
\begin{equation}
\label{equ:objective_f}
\begin{aligned}
    f(\hat{s})
    =&\sum\limits_{i;j\in{x,y,z}} \Cov^2(\hat{m}_{o}^i, m_c^j) \\
    %=&trace \left(diag \left(Vec(\mathbf{R})  \right) diag \left(Vec(\mathbf{R})  \right)   \right)\\
    =&\sum\limits_{i;j\in{x,y,z}} (\hat{s}\Cov(m_d^i,m_c^j) + \Cov(m_c^i, m_c^j))^2\\
    %=&\sum\limits_{i;j\in{x,y,z}} (\hat{s}E[\tilde{\mathbf{m}}_d^i\tilde{\mathbf{m}}_c^j ] + E[\tilde{\mathbf{m}}_c^j\tilde{\mathbf{m}}_c^j])^2\\
    %=&A\hat{s}^2+B\hat{s}+ \cdots
    =& \left( \sum\limits_{i;j\in{x,y,z}} \Cov^2(m_d^i,m_c^j) \right)  \hat{s}^2\\
     &+\left( \sum\limits_{i;j\in{x,y,z}} 2\Cov(m_d^i,m_c^j)\Cov(m_c^i, m_c^j)  \right)\hat{s}\\
     &+ \sum\limits_{i;j\in{x,y,z}} \Cov^2(m_c^i,m_c^j),
\end{aligned}
\end{equation}
\normalsize
which is a quadratic function in terms of $\hat{s}$, thus the optimal scale estimation 
could be represented in a closed form solution through minimizing $f(\hat{s})$:

\small
\begin{equation}
\label{equ:objective_s}
\begin{aligned}
    \hat{s}^*= -\frac{\sum\limits_{i;j\in{x,y,z}} \Cov( m_d^i,m_c^j )\Cov(m_c^i, m_c^j)}{\sum\limits_{i;j\in{x,y,z}} \Cov^2(m_d^i,m_c^j)}.
\end{aligned}
\end{equation}
\normalsize

\subsection{Observability condition derivation}
One key point we want to highlight is that the scale estimator has degenerated cases 
using a monocular sensor scheme. In other words, our estimator will not work
for all camera and object motion combinations.
Take the
case that the camera is almost static during the data accumulation for example, the optimal
scale estimation result $\hat{\mathbf{s}}$ will be tending to zero,
which is definitely not correct. This is because the observed compound object motion in the camera frame
that combines object motion and camera motion degenerates to object motion only, and the
constraint derived from the motion decomposition process doesn't hold anymore.

In addition, note that all the computation is based on the motion independence assumption, 
thus another possible degenerated case is that the camera moves exactly as how the object moves, resulting
in an invariable observation of the object in the camera frame.
Although this case is statistically of very low probability, the corresponding optimal scale
value $\hat{\mathbf{s}}$ can be arbitrary numbers.
On the other hand, however, it is also almost impossible that 
the object motion and the camera motion are strictly uncorrelated in a limited sample duration time.
Thus an error term $\Delta s$ exists in practical analysis within a specific observation duration.
%the objective function could be visualized as Fig. \ref{fig:objective_f}.
%\begin{figure}[t]
%\begin{center}
%  \includegraphics[width=0.5\columnwidth]{fig/objective_function.pdf}
%\end{center}
%\caption{Objective function in terms of estimated scale.}
%\label{fig:objective_f}
%\end{figure}
Nevertheless,
the estimation is still valid if the error term is limited to an acceptable range, 
and we call the corresponding condition as the observability condition. And our intrinsic scale 
estimator do not have to be running all the time since what we estimate is the intrinsic
scale of the tracked object instead of object depths of the input sequential images.
Estimating the intrinsic scale once in a while when the observability condition is 
satisfied is totally acceptable because of the region-based BA in our system.

In fact, the observability condition
is more important than the closed form scale estimation to some extent, since it determines
if we can accept the scale estimation or not.
One intuitive 
condition is that the objective function value with the optimal scale should be close to zero, 
but intuitive way may miss some degenerated cases, so a better
way is to study the observability condition through error term analysis (substitute Equation \ref{equ:object_motion_world2} and \ref{equ:cov_oc_particular} into Equation \ref{equ:objective_s}):

\small
\begin{equation}
\label{equ:objective_ds}
\begin{aligned}
    \frac{\Delta {s}^*}{s}=
    \frac{  \sum\limits_{i;j\in{x,y,z}} \Cov( m_o^i,m_c^j )( \Cov( m_c^i,m_c^j ) - \Cov(m_o^i, m_c^j))   }{\sum\limits_{i;j\in{x,y,z}} {(\Cov(m_c^i,m_c^j) - \Cov(m_o^i, m_c^j))}^2}.
\end{aligned}
\end{equation}
\normalsize
The desired scale estimation should satisfy $\frac{\Delta {s}^*}{s} \to 0$, 
and we summarize it as:

\small
\begin{equation}
\label{equ: requirement41}
\begin{aligned}
&\sum\limits_{i;j\in{x,y,z}}  \Cov^2(m_o^i, m_c^j) \to 0,\\
&\frac{1}{\sum\limits_{i;j\in{x,y,z}}  \Cov^2(m_c^i, m_c^j)} \to 0,
\end{aligned}
\end{equation}
\normalsize
where the first condition just meets the intuitive condition if we use the reconstructed object motion with
the optimal scale ($\hat{\mathbf{m}}_{o}(\hat{s}^*)$) instead of the real object motion ($\mathbf{m}_o$), 
so it's a necessary condition. 
The second condition is totally under control on the other hand, since although we cannot
control the camera motion, but we can at least
observe and analyze the camera motion to decide if the scale estimation is acceptable or not.

In practice, we use the threshold-based criteria instead:

\small
\begin{equation}
\label{equ: requirement42}
\begin{aligned}
&\sum\limits_{i;j\in{x,y,z}}  \Cov^2(m_{\hat{o}}^i, m_c^j) <= \epsilon_{t_1},\\
&\sum\limits_{i;j\in{x,y,z}}  \Cov^2(m_c^i, m_c^j) >= \rho_{t_1}.
\end{aligned}
\end{equation}
\normalsize

Also, for the numerical stability of Equation \ref{equ:objective_s}:\\
\begin{equation}
\label{equ: requirement_numerical}
\begin{aligned}
    \sum\limits_{i;j\in{x,y,z}} \Cov^2(m_d^i,m_c^j) >= \rho_{t_2}.\\
\end{aligned}
\end{equation}

So far, we can derive the complete observability condition as follows:
\begin{equation}
\label{equ:requirement_all}
\begin{aligned}
    &\rn{1}~&f(\hat{s}^*)<=\epsilon_{t_1}\\
    &\rn{2}~&\sum\limits_{i;j\in{x,y,z}} \Cov^2(m_c^i, m_c^j) >=  \rho_{t_1}\\
    &\rn{3}~&\sum\limits_{i;j\in{x,y,z}} \Cov^2(m_d^i, m_c^j) >=  \rho_{t_2},
\end{aligned}
\end{equation}
where $\epsilon_{t_1}$ is a threshold with small positive value, $\rho_{t_1}$ and $\rho_{t_2}$
are two thresholds with large enough positive values. 
The first two subconditions are used to satisfy the estimation accuracy requirement, and the last 
subcondition is for numerical stability consideration.
All the thresholds are set 
empirically in implementation. The more strict the condition is, the more accurate and 
robust the metric scale estimator will be, but with lower scale acceptance rate.
Overall, the observability condition makes our system robust against multiple
degenerated cases that may occur in the tracking process.

Given the estimated metric scale with the observability condition satisfied, 
we can first recover the object position in the camera frame as:
\begin{equation}
\label{equ:object_pose_camera}
\begin{aligned}
    \hat{\mathbf{p}}_{o_t}^{c_t}&=\hat{s}\cdot\hat{\bar{\mathbf{p}}}_{o_t}^{c_t},~~~~~~~~
\end{aligned}
\end{equation}
then we can recover the object poses in the world frame:
\begin{equation}
\label{equ:object_pose_world}
\begin{aligned}
    \hat{\mathbf{R}}_{o_t}^w&=\hat{\mathbf{R}}_{c_t}^w\hat{\mathbf{R}}_{o_t}^{c_t},\\
    \hat{\mathbf{p}}_{o_t}^w&=\hat{s}\cdot\hat{\mathbf{R}}_{c_t}^w\hat{\bar{\mathbf{p}}}_{o_t}^{c_t} + \hat{\mathbf{p}}_{c_t}^w.
\end{aligned}
\end{equation}
It is evident that the final object localization performance relies on VINS ($\hat{\mathbf{p}}_{c_t}^w$ and $\hat{\mathbf{R}}_{c_t}^w$), up-to-scale 
$3$D tracking ($\hat{\bar{\mathbf{p}}}_{o_t}^{c_t}$ and $\hat{\mathbf{R}}_{o_t}^{c_t}$) and metric scale estimation results ($\hat{s}$).

\begin{figure}[t]
\begin{center}
    \subfigure[The sensor suite]{\includegraphics[width=0.312\columnwidth]{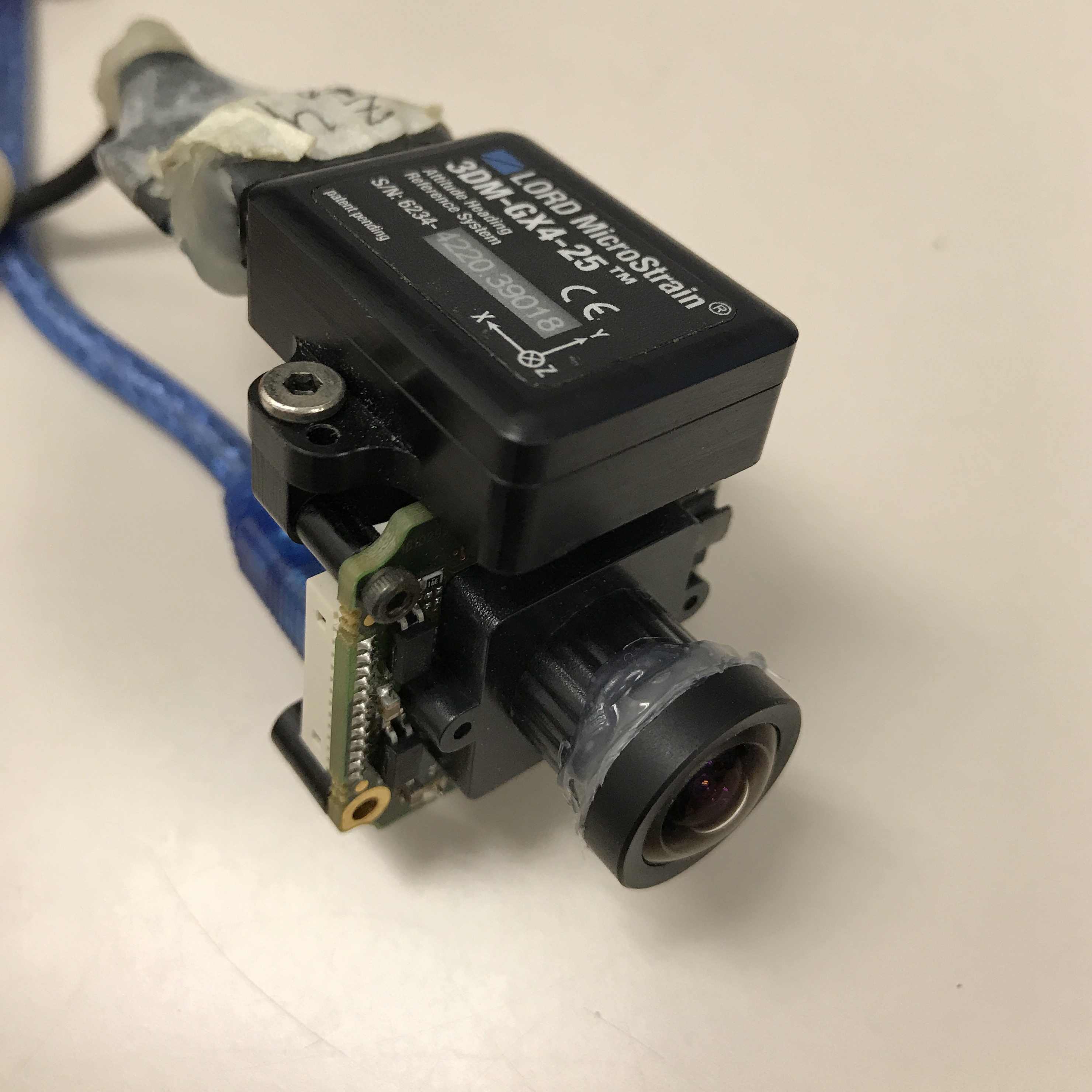}}
    \subfigure[The objects for tracking]{\includegraphics[width=0.635\columnwidth]{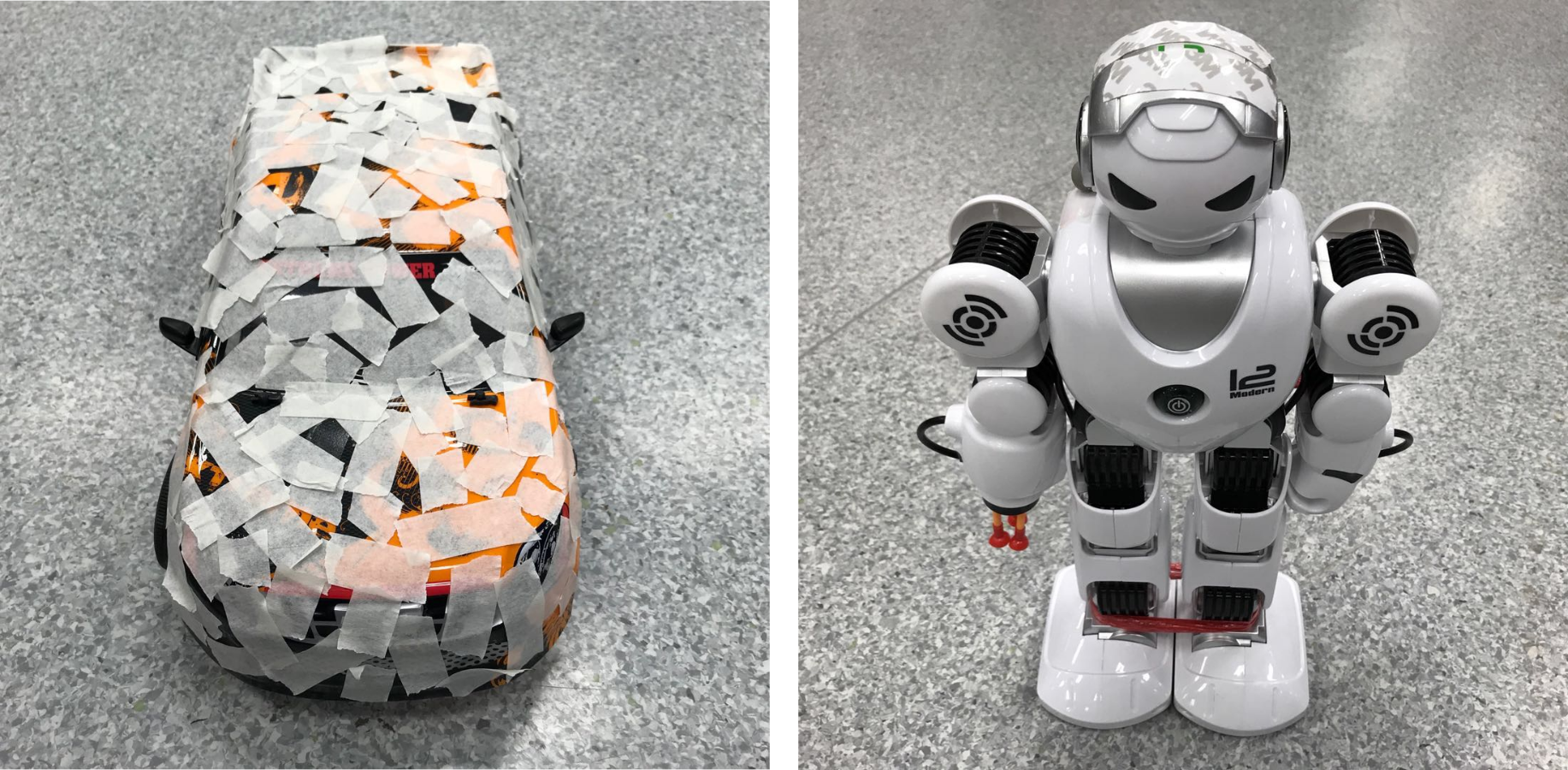}}  
\end{center}
  \caption{The minimal sensor suite and irregular objects we use for experiments. Our method do not need any prior knowledge, special markers or tags of the tracking object.}
\label{fig:hardware}
\vspace{-5mm}
\end{figure}

\section{Experimental Results}
\label{sec:experiments}
\subsection{Implementation details}
Our monocular sensor suite (Fig.~\ref{fig:hardware}(a))
consists of an mvBlueFOX-MLC202bG grayscale camera
with a wide-angle lens that captures $1280$ by $960$ images at $24$ Hz and a Microstrain 3DM-GX4 IMU that runs at $500$ Hz. 
Both VINS and $2$D tracker work on the downsampled images of $640$ by $480$ resolution. 
We use CMT tracker \cite{nebehay2015clustering} as the $2$D tracker in our implementation and 
manually draw the object area for $2$D tracking initialization.
While the cropped images according to $2$D tracking results are used for region-based 
bundle adjustment ($N=20$ in sliding window). 
For robust region-based BA estimation, the object area extracted from the original image cannot
be too small, which means the object cannot be too far away from the camera. In practice, 
we notice that at least $50$ features should be extracted from the object region if we 
use feature-based tracking frontend. The number of observations we use for metric scale 
estimation is $200$ ($N_o=200$). The motion feature extracted is velocity ($n=1$ in Equation \ref{equ:RV}).
%We use the Linux OS with ROS as the robotics middleware.
The whole system runs on a desktop computer in real time.

The goal of the experiments is to validate the recovered object pose in the world 
frame ($\mathbf{R}_o^w$ and $\mathbf{p}_o^w$) 
by involving the camera pose in the 
world frame and the estimated metric scale. Since there are no similar formulations shown in 
relevant works,
%model-based method can estimate the metric scale with only one
%observation while our method needs to accumulate a period of observations for 
%metric scale estimation; trajectory reconstruction estimates point depth for 
%each observation while our method estimates the intrinsic scale together with the rotation 
%of a rigid object.
we evaluate our estimation
accuracy with a comparison to a motion capture system (OptiTrack) as groundtruth.
And we use several general objects as the tracking targets. Two sample objects 
are shown in Fig.~\ref{fig:hardware}(b), where the tapes attached on the objects
are used to cover the excessively shining parts that may deactivate the motion
capture system.

\subsection{$6$-DoF dynamic object pose estimation in world frame}
Given the estimated metric scale, the $3D$ tracking results could be projected to
the world frame using Equation \ref{equ:object_pose_world}. 
For trial $1$, a manually controlled toy robot is adopted as the tracking target.
While for trial $2$, a remote controlled racing car is used as the tracking target, which 
is controlled to run from ground to a higher platform with directions changing all the time. More details
and AR demonstration will be shown in the attached video.

\begin{figure}[t]
\begin{center}
    \subfigure[position]{\includegraphics[width=0.6\columnwidth]{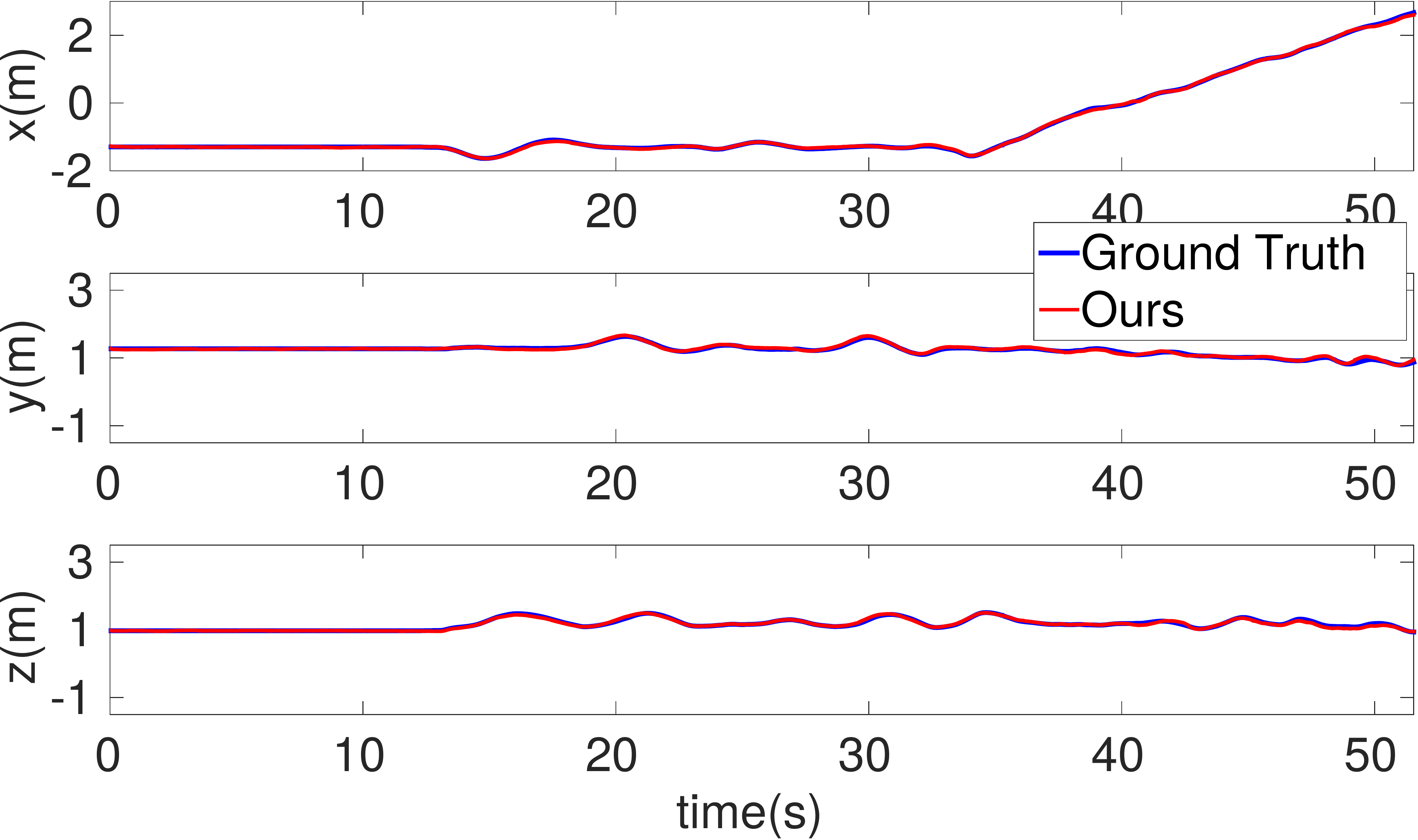}}
    \subfigure[orientation]{\includegraphics[width=0.6\columnwidth]{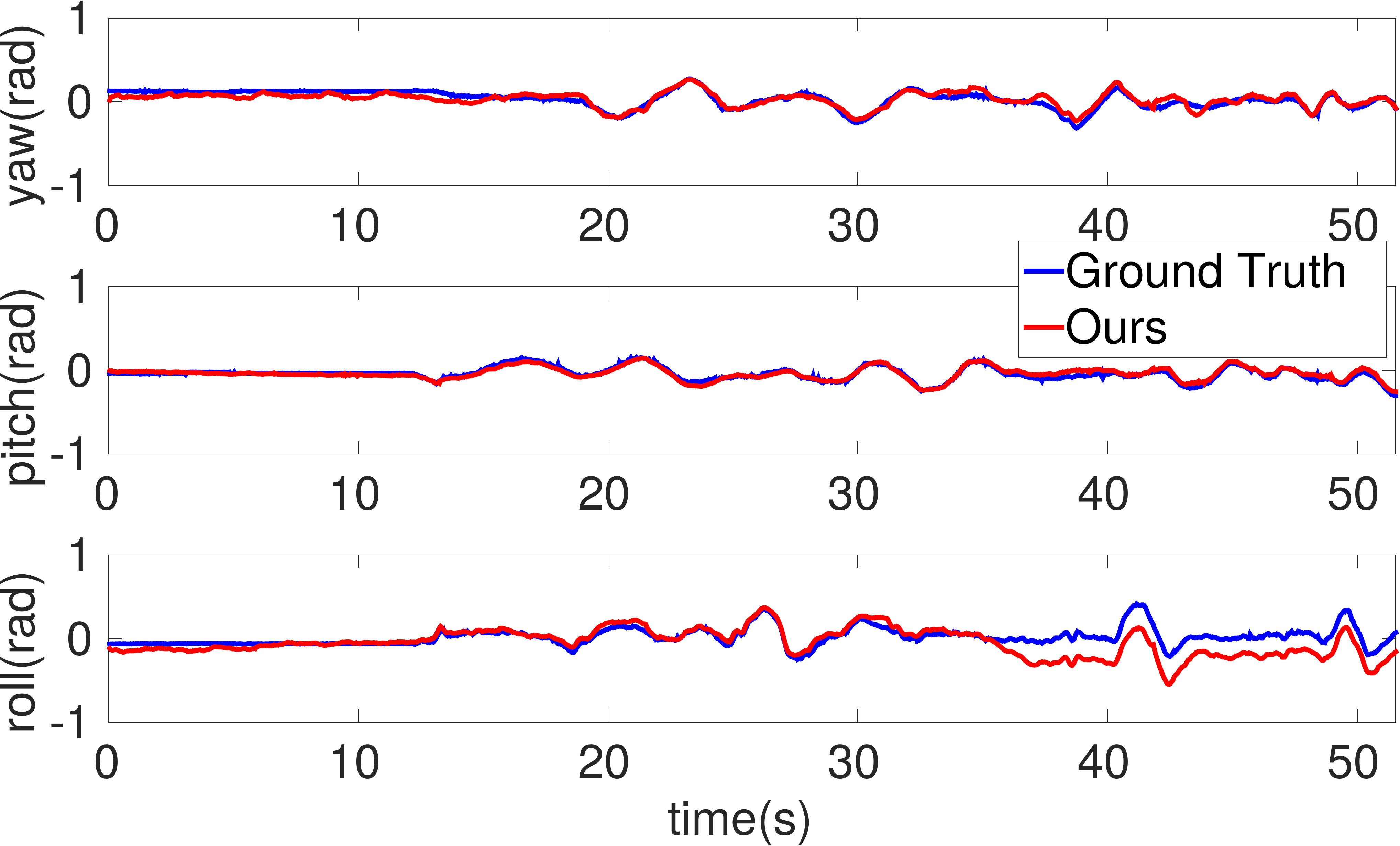}}
\end{center}
\caption{The comparison of $6$-DoF object pose estimation (Trial $1$: tracking a manually controlled toy robot).}
\label{fig:vicon_mono_trial1}
\end{figure}

\begin{figure}[t]
\begin{center}
    \subfigure[position]{\includegraphics[width=0.6\columnwidth]{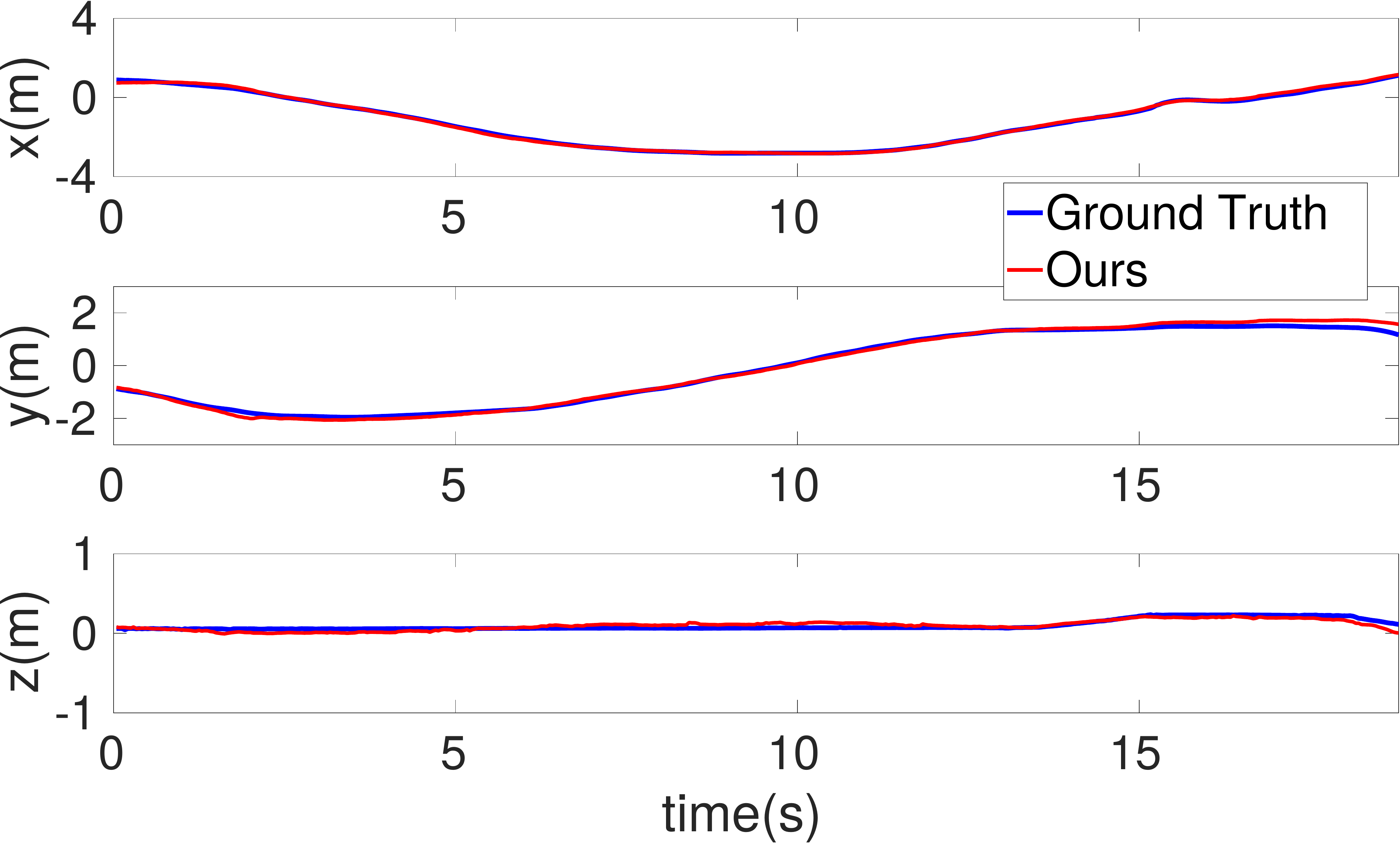}}
    \subfigure[orientation]{\includegraphics[width=0.6\columnwidth]{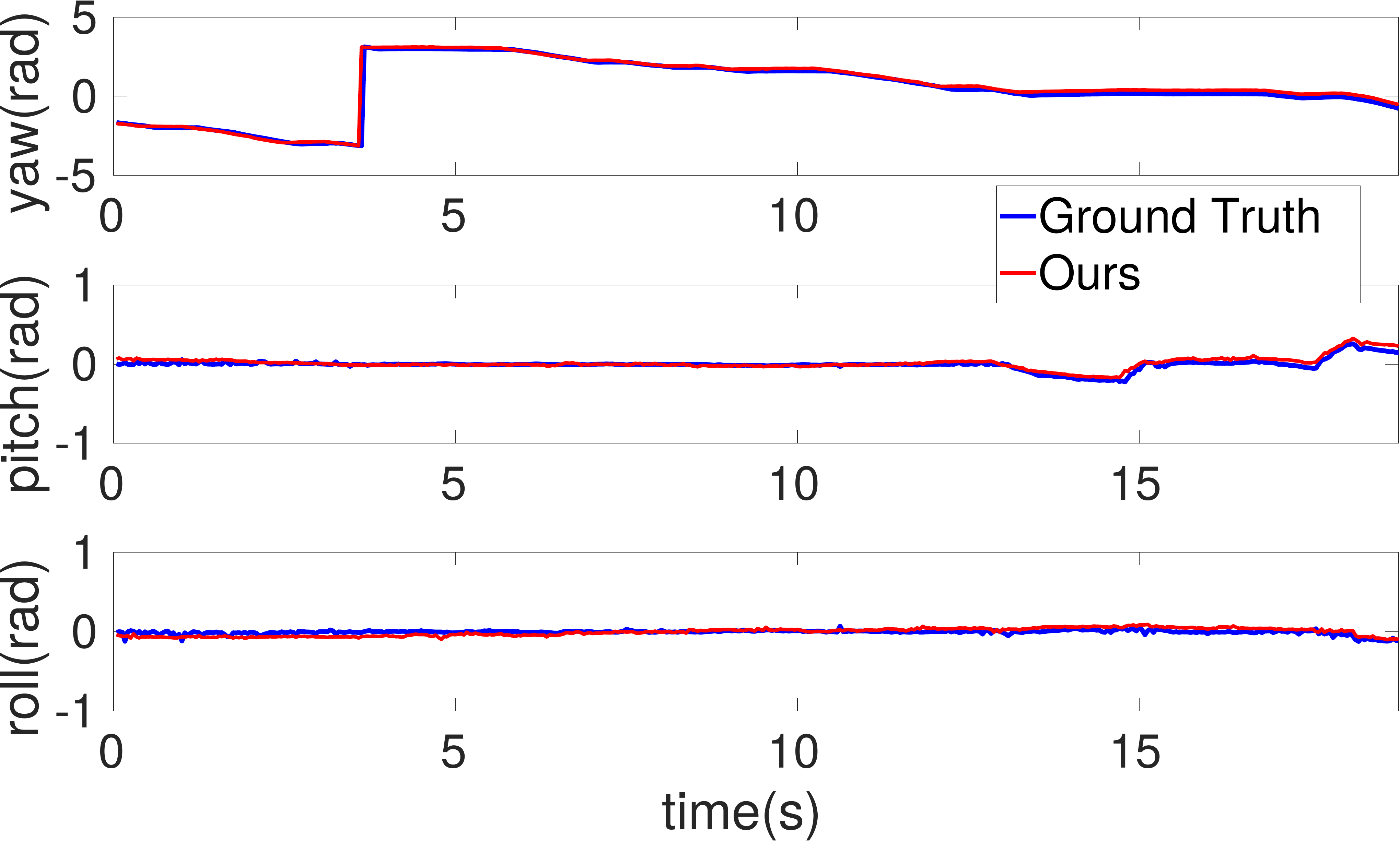}}
\end{center}
\caption{The comparison of $6$-DoF object pose estimation (Trial $2$: tracking a remote controlled racing car).}
\label{fig:vicon_mono_trial2}
\vspace{-5mm}
\end{figure}

\begin{table*}
\caption{Degenerated cases}
\label{table:cases}
\centering
\begin{tabular}[h]{cccc}
    \toprule
    Cases & true scale $s$ & sample estimations $\hat{s}$ & observability subcondition \\ \midrule
    $\mathrm{C}_1$ & 0.43 & (-5.99~-4.62~-2.71~-1.13~16.87~35.85~12.27~17.19~-0.38~-0.26) & (\rn{1}) not satisfied \\ \midrule
    $\mathrm{C}_2$ & 0.43 & (0.047~-0.008~-0.039~-0.041~-0.007~-0.031~-0.067~0.017~0.008~0.009) &  (\rn{2}) not satisfied \\ \midrule
    $\mathrm{C}_3$ & 0.43 & (-0.085~-0.088~-0.0085~-0.078~-0.077~-0.085~-0.078~-0.085~-0.078~-0.076) & (\rn{2}) not satisfied \\ \midrule
    $\mathrm{C}_4$ & 0.43 & (21.81~20.82~19.76~18.82~27.38~25.14~21.57~22.77~21.57~24.15) & (\rn{3}) not satisfied \\ \midrule
\bottomrule
  \end{tabular}
\end{table*}

Fig. \ref{fig:vicon_mono_trial1} and Fig. \ref{fig:vicon_mono_trial2} show part of the position and 
orientation comparison between the estimation and groundtruth,
with a standard deviation of $\{0.0486, 0.0271, 0.1245\}$ ($\mathrm{rad}$) in yaw, pitch and roll, 
a standard deviation of $\{0.0218, 0.0310, 0.0344\}$ ($\mathrm{m}$) in the $x$, $y$ and $z$ positions
of trial $1$,
and a standard deviation of $\{0.0739, 0.0254, 0.0239\}$ ($\mathrm{rad}$) in yaw, pitch and roll, 
a standard deviation of $\{0.0505, 0.1169, 0.0397\}$ ($\mathrm{m}$) in the $x$, $y$ and $z$ positions
of trial $2$.
Generally, the experiment shows acceptable performance of our $6$-DoF object tracking system. 
%since the 
%metric accuracy of the estimated object pose in world frame may not be the first concern for some
%applications. 
We can also notice the orientation drift of trial $1$ and translation drift of trial $2$ caused by the 
up-to-scale $3$D tracker module, which could 
be corrected by visual relocalization or loop closure if a loop is detected.

\subsection{Degenerated cases}
We further test three typical degenerated cases that will generate inaccurate scale estimation without checking
the observability condition:
the camera moves exactly as how the car moves in terms of translation ($\mathrm{C}_1$); the camera is inactive, 
static ($\mathrm{C}_2$), constant velocity ($\mathrm{C}_3$); and 
static object observation in the camera frame although both the car and the camera are moving ($\mathrm{C}_4$),
which correspond to the three observability
subconditions discussed in Equation \ref{equ:requirement_all} respectively. 
And all these degenerated cases could be avoided by checking
the observability condition.
The true scale, sample scale estimations under these cases, and corresponding subconditions are listed in Table \ref{table:cases}.
According to Equation \ref{equ:objective_ds}, the scale estimations may have
large estimation errors for case $C_1$ ($\Cov(m_o^i,m_c^j)~\textbf{not} \to 0$). According to Equation \ref{equ:objective_s}, the scale 
estimations tend to be approaching to zero for case $C_2$ and case $C_3$ ($\Cov(m_c^i,m_c^j) \to 0$).
And according to Equation \ref{equ:objective_s}, the scale estimations tend to have large absolute values for case $C_4$
($\Cov(m_d^i,m_c^j) \to 0$). Fortunately, all the degenerated cases could be avoided by checking
the observability condition during the tracking process.

\section{Conclusion and Future Work}
\label{sec:conclusion}
In this paper, we propose a new understanding of the $6$-DoF object tracking problem using a monocular sensor
suite with visual-inertial fusion. 
We have proven the feasibility of the proposed $3$D tracking system with observability condition analysis.
The experimental results validate our idea and show an acceptable estimation accuracy 
of the proposed method.
At the same time, each module in our system has upgrading potential. 
For example, the region-based bundle adjustment implemented
in this paper is purely a BA without relocalization, loop-closure which are widely used 
in the visual odometry community. Thanks to the building block design of our system, fortunately, each module can be 
conveniently replaced, and the whole system will be boosted accordingly.

%\section{Video Supplement}

\bibliographystyle{IEEEtran}
\bibliography{iros_tracking_final.bib}

\begin{thebibliography}{10}
\providecommand{\url}[1]{#1}
\csname url@samestyle\endcsname
\providecommand{\newblock}{\relax}
\providecommand{\bibinfo}[2]{#2}
\providecommand{\BIBentrySTDinterwordspacing}{\spaceskip=0pt\relax}
\providecommand{\BIBentryALTinterwordstretchfactor}{4}
\providecommand{\BIBentryALTinterwordspacing}{\spaceskip=\fontdimen2\font plus
\BIBentryALTinterwordstretchfactor\fontdimen3\font minus
  \fontdimen4\font\relax}
\providecommand{\BIBforeignlanguage}[2]{{%
\expandafter\ifx\csname l@#1\endcsname\relax
\typeout{** WARNING: IEEEtran.bst: No hyphenation pattern has been}%
\typeout{** loaded for the language `#1'. Using the pattern for}%
\typeout{** the default language instead.}%
\else
\language=\csname l@#1\endcsname
\fi
#2}}
\providecommand{\BIBdecl}{\relax}
\BIBdecl

\bibitem{lin2017autonomous}
Y.~Lin, F.~Gao, T.~Qin, W.~Gao, T.~Liu, W.~Wu, Z.~Yang, and S.~Shen,
  ``Autonomous aerial navigation using monocular visual-inertial fusion,''
  \emph{Journal of Field Robotics}, 2017.

\bibitem{yang2017real}
Z.~Yang, F.~Gao, and S.~Shen, ``Real-time monocular dense mapping on aerial
  robots using visual-inertial fusion,'' in \emph{Robotics and Automation
  (ICRA), 2017 IEEE International Conference on}.\hskip 1em plus 0.5em minus
  0.4em\relax IEEE, 2017, pp. 4552--4559.

\bibitem{aldoma2013multimodal}
A.~Aldoma, F.~Tombari, J.~Prankl, A.~Richtsfeld, L.~Di~Stefano, and M.~Vincze,
  ``Multimodal cue integration through hypotheses verification for rgb-d object
  recognition and 6dof pose estimation,'' in \emph{Robotics and Automation
  (ICRA), 2013 IEEE International Conference on}.\hskip 1em plus 0.5em minus
  0.4em\relax IEEE, 2013, pp. 2104--2111.

\bibitem{wang20173d}
M.-S. Wang \emph{et~al.}, ``3d object pose estimation using stereo vision for
  object manipulation system,'' in \emph{Applied System Innovation (ICASI),
  2017 International Conference on}.\hskip 1em plus 0.5em minus 0.4em\relax
  IEEE, 2017, pp. 1532--1535.

\bibitem{rozantsev2016flight}
A.~Rozantsev, S.~N. Sinha, D.~Dey, and P.~Fua, ``Flight dynamics-based recovery
  of a uav trajectory using ground cameras,'' \emph{arXiv preprint
  arXiv:1612.00192}, 2016.

\bibitem{vo2016spatiotemporal}
M.~Vo, S.~G. Narasimhan, and Y.~Sheikh, ``Spatiotemporal bundle adjustment for
  dynamic 3d reconstruction,'' in \emph{Proceedings of the IEEE Conference on
  Computer Vision and Pattern Recognition}, 2016, pp. 1710--1718.

\bibitem{garrido2014automatic}
S.~Garrido-Jurado, R.~Mu{\~n}oz-Salinas, F.~J. Madrid-Cuevas, and M.~J.
  Mar{\'\i}n-Jim{\'e}nez, ``Automatic generation and detection of highly
  reliable fiducial markers under occlusion,'' \emph{Pattern Recognition},
  vol.~47, no.~6, pp. 2280--2292, 2014.

\bibitem{qiu2017model}
K.~Qiu, T.~Liu, and S.~Shen, ``Model-based global localization for aerial
  robots using edge alignment,'' \emph{IEEE Robotics and Automation Letters},
  vol.~2, no.~3, pp. 1256--1263, 2017.

\bibitem{qiu2017model2}
K.~Qiu and S.~Shen, ``Model-aided monocular visual-inertial state estimation
  and dense mapping,'' in \emph{Intelligent Robots and Systems (IROS), 2017
  IEEE/RSJ International Conference on}.\hskip 1em plus 0.5em minus 0.4em\relax
  IEEE, 2017, pp. 1783--1789.

\bibitem{Aruco2014}
S.~Garrido-Jurado, R.~M. noz Salinas, F.~Madrid-Cuevas, and
  M.~Mar\'in-Jim\'enez, ``Automatic generation and detection of highly reliable
  fiducial markers under occlusion,'' \emph{Pattern Recognition}, vol.~47,
  no.~6, pp. 2280 -- 2292, 2014.

\bibitem{leutenegger2011brisk}
S.~Leutenegger, M.~Chli, and R.~Y. Siegwart, ``Brisk: Binary robust invariant
  scalable keypoints,'' in \emph{Computer Vision (ICCV), 2011 IEEE
  International Conference on}.\hskip 1em plus 0.5em minus 0.4em\relax IEEE,
  2011, pp. 2548--2555.

\bibitem{brachmann2016uncertainty}
E.~Brachmann, F.~Michel, A.~Krull, M.~Ying~Yang, S.~Gumhold \emph{et~al.},
  ``Uncertainty-driven 6d pose estimation of objects and scenes from a single
  rgb image,'' in \emph{Proceedings of the IEEE Conference on Computer Vision
  and Pattern Recognition}, 2016, pp. 3364--3372.

\bibitem{choi20123d}
C.~Choi and H.~I. Christensen, ``3d textureless object detection and tracking:
  An edge-based approach,'' in \emph{Intelligent Robots and Systems (IROS),
  2012 IEEE/RSJ International Conference on}.\hskip 1em plus 0.5em minus
  0.4em\relax IEEE, 2012, pp. 3877--3884.

\bibitem{zeng2017multi}
A.~Zeng, K.-T. Yu, S.~Song, D.~Suo, E.~Walker, A.~Rodriguez, and J.~Xiao,
  ``Multi-view self-supervised deep learning for 6d pose estimation in the
  amazon picking challenge,'' in \emph{Robotics and Automation (ICRA), 2017
  IEEE International Conference on}.\hskip 1em plus 0.5em minus 0.4em\relax
  IEEE, 2017, pp. 1386--1383.

\bibitem{pavlakos20176}
G.~Pavlakos, X.~Zhou, A.~Chan, K.~G. Derpanis, and K.~Daniilidis, ``6-dof
  object pose from semantic keypoints,'' \emph{arXiv preprint
  arXiv:1703.04670}, 2017.

\bibitem{yang2016monocular}
Z.~Yang and S.~Shen, ``Monocular visual-inertial state estimation with online
  initialization and camera-imu extrinsic calibration,'' \emph{IEEE
  Transactions on Automation Science and Engineering}, no.~99, p.~1, 2016.

\bibitem{QinShen17}
T.~Qin and S.~Shen, ``Robust initialization of monocular visual-inertial
  estimation on aerial robots.'' in \emph{Proc. of the {IEEE/RSJ} Intl. Conf.
  on Intell. Robots and Syst.}, Vancouver, Canada, 2017, accepted.

\bibitem{nebehay2015clustering}
G.~Nebehay and R.~Pflugfelder, ``Clustering of static-adaptive correspondences
  for deformable object tracking,'' in \emph{Proceedings of the IEEE Conference
  on Computer Vision and Pattern Recognition}, 2015, pp. 2784--2791.

\bibitem{hare2016struck}
S.~Hare, S.~Golodetz, A.~Saffari, V.~Vineet, M.-M. Cheng, S.~L. Hicks, and
  P.~H. Torr, ``Struck: Structured output tracking with kernels,'' \emph{IEEE
  transactions on pattern analysis and machine intelligence}, vol.~38, no.~10,
  pp. 2096--2109, 2016.

\bibitem{hooper1959simultaneous}
J.~W. Hooper, ``Simultaneous equations and canonical correlation theory,''
  \emph{Econometrica: Journal of the Econometric Society}, pp. 245--256, 1959.

\end{thebibliography}
\end{document}